\documentclass[conference]{IEEEtran}

\usepackage{amsmath,amssymb}
\usepackage{graphicx}
\usepackage{booktabs}
\usepackage{array}
\usepackage{url}
\usepackage[caption=false,font=footnotesize]{subfig}
\usepackage{dblfloatfix}
\usepackage[section]{placeins}

\DeclareGraphicsExtensions{.pdf,.png,.jpg}

\ifdefined\pdfminorversion\pdfminorversion=7\fi

\title{Large Language Model based Interactive Decision-Making for Autonomous Driving}
\author{Xinwei DONG\IEEEauthorrefmark{1}, Jiyang LI\IEEEauthorrefmark{2}, Jiabin XIE\IEEEauthorrefmark{1}, Yang YI\IEEEauthorrefmark{3}, Tianshang JIA\IEEEauthorrefmark{1}, Shiyu FANG\IEEEauthorrefmark{1}, Ye TIAN\IEEEauthorrefmark{1}, Peng HANG\IEEEauthorrefmark{1}\thanks{Corresponding author: hangpeng@tongji.edu.cn}\\
\IEEEauthorrefmark{1}College of Transportation, Tongji University, Shanghai 201804, China\\ \IEEEauthorrefmark{2}School of Automotive Studies, Tongji University, Shanghai 201804, China\\ \IEEEauthorrefmark{3}School of Computer Science and Technology, Tongji University, Shanghai 201804, China}

\begin{document}
\maketitle

\begin{abstract}
In high-conflict mixed-traffic scenarios involving human-driven and autonomous vehicles, most existing autonomous driving systems default to overly conservative behaviors, lack proactive interaction, and consequently suffer from limited public acceptance. To mitigate intent misunderstandings and decision failures, we present a Large Language Model based interactive decision-making framework that augments scene understanding and intent-aware interaction to jointly improve safety and efficiency. The approach uses Object-Process Methodology to semantically model complex multi-vehicle scenes, abstracting low-level perceptual data into objects, processes, and relations, thereby streamlining reasoning over latent causal structure. Building on this representation, the Large Language Model parses both explicit and implicit intents of surrounding agents and, under jointly enforced safety and efficiency constraints, selects candidate maneuvers. We further generate perturbed trajectory candidates via Monte Carlo sampling and evaluate them to obtain an optimized executable trajectory. To foster transparency and coordination with nearby road users, the final decision is translated by the Large Language Model into concise natural-language messages and broadcast through an external Human--Machine Interface, completing a closed loop from scene understanding to action to language. Experiments in a cluster driving simulator demonstrate that the proposed method outperforms traditional baselines across safety, comfort, and efficiency metrics, while a Turing-test-style evaluation indicates a high degree of human-likeness in decision making. Besides, these results suggest that coupling semantic scene abstraction with Large Language Model mediated intent reasoning and language-based eHMI communication offers a practical pathway toward interactive, trustworthy autonomous driving in dense mixed traffic.
\end{abstract}

\renewcommand{\IEEEkeywordsname}{Keywords}

\section{Introduction}

In February 2020, eleven Chinese ministries, including the National Development and Reform Commission, jointly issued the Strategy for Innovation and Development of Intelligent Vehicles, which explicitly outlined that by 2025, China aims to achieve large-scale production of conditionally autonomous intelligent vehicles and regional coverage of vehicle wireless communication networks (e.g., LTE-V2X). Consequently, the penetration rate of Connected Automated Vehicles (CAVs) is expected to increase steadily (Olayode et al., 2023; Beijing Innovation Center for Intelligent Vehicles and Mobility Industry, 2022).

Despite this progress, autonomous driving systems continue to exhibit insufficient decision-making intelligence in mixed human--autonomous traffic environments, particularly when confronted with complex, high-conflict, and stochastic interaction scenarios. Representative incidents---including multiple Cruise autonomous vehicles becoming immobilized at an intersection in San Francisco in 2022 and decision failures of Baidu Apollo autonomous taxis in Wuhan in 2024---highlight the vulnerability of current systems in such settings. Accident statistics further show that more than 80\% of autonomous driving failures occur in complex conflict zones such as intersections and merging areas (Beijing Innovation Center for Intelligent Vehicles and Mobility Industry, 2022). These failures commonly stem from ambiguous right-of-way and dense multi-vehicle interactions, which often trigger overly conservative or unclear behaviors by autonomous vehicles. Such behaviors deviate from human driving expectations, confuse surrounding drivers, and may escalate into secondary conflicts or accidents (Pagale et al., 2024). At their core, these issues reflect the limited ability of existing autonomous systems to proactively understand and interact with human road users, particularly in risk identification and conflict anticipation stages (Ge et al., 2023).

For the foreseeable future, road traffic will remain a mixed driving environment where human-driven vehicles (HDVs) and CAVs coexist. In this context, autonomous vehicles must improve not only decision optimality but also interactive intelligence, namely the capability to anticipate, negotiate, and coordinate with human drivers. Existing approaches to autonomous driving decision-making can be broadly classified into three categories: rule-based methods, game-theoretic methods, and machine learning-based methods.

Rule-based approaches rely on predefined rules or handcrafted behavioral models and typically yield safe and predictable behavior in routine scenarios. A representative example is the ``first-come, first-served'' strategy, in which right-of-way is assigned based on arrival order. This principle has been extended into reservation-based and search-based coordination frameworks---such as Monte Carlo Tree Search, Depth-First Search, and Minimum Clique Coverage---that optimize vehicle passage by enforcing exclusive occupancy of conflict zones (Xu et al., 2019). While effective in structured environments or traffic streams dominated by CAVs, these methods struggle in mixed traffic due to the diversity and unpredictability of human driving behavior, often forcing CAVs into passive responses that undermine efficiency and interaction quality.

Game-theoretic methods explicitly model traffic as a strategic multi-agent system and derive interactive decisions through equilibrium concepts, enabling a certain degree of proactive behavior. By balancing competition and cooperation, these methods allow CAVs to adjust strategies in response to human driving behaviors (Schwarting et al., 2019). Typical formulations include Stackelberg games (Rahmati et al., 2022), cooperative games (Hang et al., 2021), and hierarchical games with K-value--based right-of-way allocation (Lu et al., 2023; Ji et al., 2023). However, in real mixed traffic, limited observability and heterogeneous driving styles make it difficult to accurately model opponents' intents and payoffs. Moreover, the computational burden of solving multi-agent games grows rapidly with the number of participants, posing significant challenges for real-time deployment.

Machine learning-based methods, particularly imitation learning and reinforcement learning, which has been extensively reviewed in recent surveys on autonomous driving decision-making and planning (Hu et al., 2025), offer a data-driven alternative by learning driving policies from demonstrations or interaction experience. Imitation learning can capture latent social norms from historical data and performs well in common scenarios (Munigety, 2018), but generalizes poorly to long-tail or unfamiliar situations. Reinforcement learning improves adaptability including end-to-end deep reinforcement learning approaches (Peng et al., 2023) through trial-and-error optimization; for example, Liu et al. proposed a MADDPG-based framework with an attention mechanism to better model CAV--HDV interactions and improve intersection efficiency (Liu et al., 2025). Nevertheless, learning-based approaches still face persistent challenges in multi-task transfer, scenario generalization, and most critically interpretability, which limits trust and hampers explicit intent negotiation in highly interactive mixed-traffic scenarios.

Human--machine interaction (HMI), especially external human--machine interface (eHMI), has been recognized as an important complementary mechanism for improving mutual understanding between autonomous vehicles and surrounding road users. Prior studies on both in-vehicle HMI and eHMI indicate that explicit intent communication through visual, auditory, or textual cues can reduce uncertainty, improve trust, and enhance safety in interaction-intensive scenarios (Zhang, 2020; Lee et al., 2021; Zhang et al., 2021; Gao and Martens, 2022; Faas and Baumann, 2019). Empirical investigations further confirm that well-designed eHMI signals can significantly accelerate intent inference and coordination among road users, particularly in intersection and shared-space scenarios (Brill et al., 2023). However, most existing HMI and eHMI designs remain largely signal-centric, focusing on interface presentation rather than being tightly coupled with the internal decision-making and reasoning processes of autonomous vehicles, which results in fragmented interaction rather than truly coordinated decision-making.

In recent years, large language models (LLMs) have exhibited strong capabilities in semantic understanding, contextual reasoning, and intent inference, offering new opportunities to bridge perception, decision-making, and interaction in autonomous driving systems. Representative studies have explored the use of LLMs for semantic scene interpretation, intent reasoning, and explainable decision support, including language-driven behavior reasoning frameworks (Cui et al., 2024), knowledge-augmented decision support models (Hu et al., 2025), and end-to-end vision--language driving systems (Xu et al., 2024; Yang et al., 2024), collectively demonstrating the potential of LLMs to enhance interpretability and human-likeness. However, existing LLM-based approaches largely focus on end-to-end perception--action mapping or post hoc explanation, and still face critical challenges related to real-time applicability, hallucination risks, and the lack of structured scene representations aligned with causal reasoning.

Therefore, a clear research gap remains: current autonomous driving decision-making frameworks lack an integrated mechanism that abstracts complex traffic scenes into semantically structured representations suitable for reasoning, explicitly models and reasons about multi-agent intents in mixed traffic, and closes the interaction loop by coherently linking decision-making outcomes with natural-language-based eHMI communication.

The main contributions of this work are threefold:

\begin{itemize}
\item An Object--Process Methodology (OPM)-based semantic scene modeling method that emulates human cognitive reasoning by abstracting complex mixed-traffic environments into structured object--process--relation representations, thereby facilitating interpretable and efficient LLM-based decision reasoning.

\item An LLM-driven interactive decision-making framework that explicitly integrates multi-agent intent parsing with safety and efficiency aware behavior selection. By transforming low-level perceptual data into structured object--process--relation representations, the proposed framework enables efficient and interpretable reasoning over multi-vehicle interactions. Building on this semantic foundation, the LLM performs intent-aware decision-making

\item A language-based eHMI interaction mechanism that translates autonomous driving decisions into natural language, enhancing transparency, coordination, and social acceptance, as validated through simulator-based experiments and Turing-test-style evaluations.

\end{itemize}
The remainder of this paper is structured as follows. Section 2 outlines the overall framework and methodology. Section 3 presents the Object-Process Methodology-based scene parsing approach. Section 4 details the integration of Large Language Models in the decision-making process, including intent parsing, decision tree construction and correction, and trajectory optimization. Section 5 describes the natural language interaction mechanism enabled by integrating Large Language Models with external Human-Machine Interfaces. Section 6 reports experimental validation through simulator-based comparative analysis and Turing test evaluation. Section 7 concludes the paper and discusses future research directions.

\section{Materials and methods}

To strengthen decision-making in complex mixed traffic, we integrate OPM-based scene understanding with an LLM-driven interactive decision module. The pipeline, as illustrated in Fig. 1, comprises four stages: scene understanding, behavior decision-making, trajectory planning, and intent interaction.

\begin{figure}[!t]
\centering
\includegraphics[width=\linewidth]{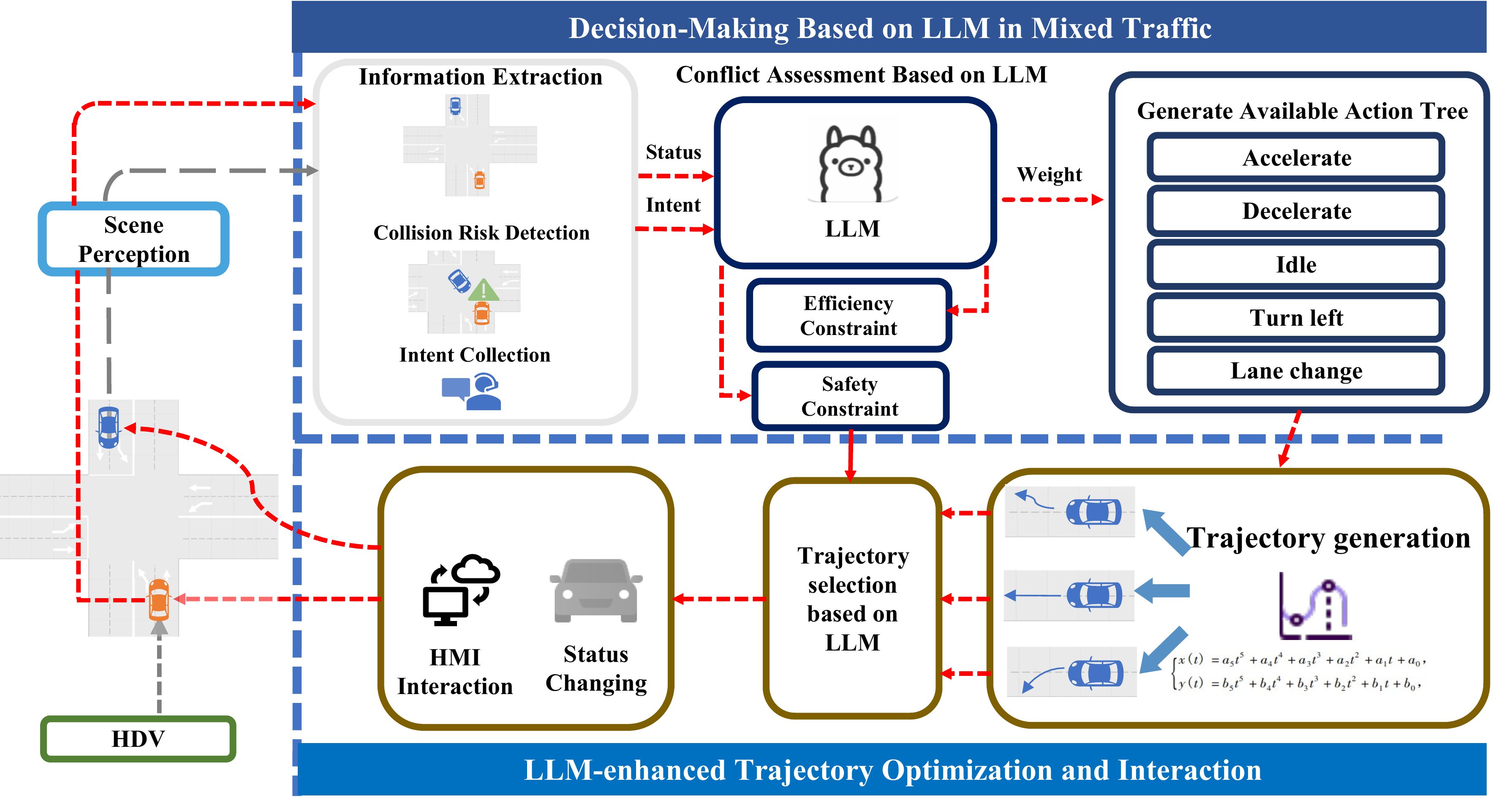}
\caption{LLM-based Interactive Autonomous Driving Framework}
\end{figure}

In the scene understanding module, the system collects real-time position and kinematic states, which can be further enhanced by cooperative perception techniques (Wang et al., 2023), then performs preliminary trajectory conflict detection. Subsequently, raw data obtained from the perception layer is transformed into a semantic structure compliant with the OPM paradigm, achieving a mapping from discrete perceptual information to a structured semantic graph. This graph comprises three core elements:

(1) Objects: Traffic participants such as vehicles and pedestrians;

(2) Processes: Current operational states and behaviors, such as lane changing, deceleration, and steering;

(3) Relations: Interaction relationships among traffic participants, including potential collisions and right-of-way conflicts.

Compared to traditional decision-making models driven by temporal features, OPM emphasizes semantic structural representation aligning closely with the language understanding paradigm of LLMs. This design significantly improves contextual modeling and the efficiency of logical reasoning.

As illustrated in Fig. 2, the OPM semantic structure enables LLMs to more clearly acquire core scene information, mitigating redundant perceptual noise and improving understanding accuracy. However, structured scene information alone is insufficient to support high-quality decision-making; the system must also conduct in-depth parsing of traffic participants\textquotesingle{} intents.

\begin{figure}[!t]
\centering
\includegraphics[width=\linewidth]{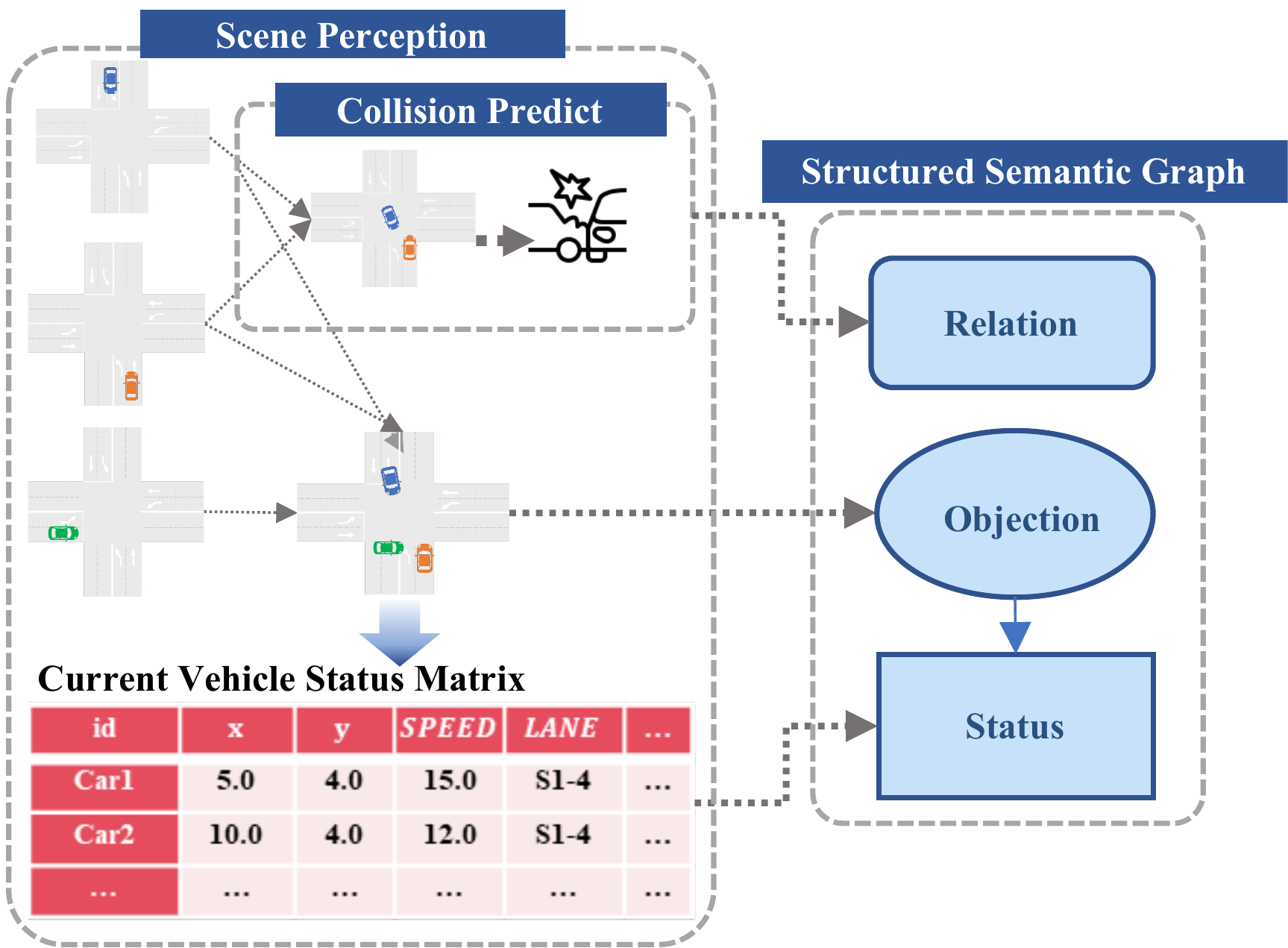}
\caption{Mapping scene raw information to structured semantic graph}
\end{figure}

For implicit intents, the system infers them by integrating historical behavioral trajectories, acceleration changes, and other relevant data. For explicit intents expressed through signals such as lights, voice, or other modalities, the LLM performs semantic parsing and decision-making. Following intent recognition, the system comprehensively considers safety and efficiency constraints to select the optimal decision trajectory from candidate strategies.

Ultimately, the selected decision is translated by the LLM into natural language descriptions and communicated to surrounding traffic participants via the eHMI. This process establishes a closed-loop workflow of perception, understanding, decision-making, and language interaction, enhancing the interpretability and social acceptance of autonomous vehicles in mixed traffic environments.

\section{Scene Parsing Based on OPM}

In complex mixed driving environments, autonomous vehicles must simultaneously comprehend the behavioral states and interaction relationships of multiple traffic participants. Directly inputting the vast amount of discrete data obtained from the perception layer into a LLM is not only computationally inefficient but also prone to inaccurate semantic understanding due to the lack of structured representation. To address this, this study adopts the OPM for semantic abstraction and pre-structured modeling of scenes. This approach constructs a semantic representation that aligns more closely with human cognitive logic, thereby enhancing the LLM's reasoning efficiency and strategy generation capabilities.

\subsection{Modeling Object Selection and Formal Representation}

Let $\mathcal{V}$ denote the set of all detected traffic participants within the perception range of the ego vehicle.

Due to perception and computational constraints, only a subset of vehicles \(\mathcal{V}_{sel} \subseteq \mathcal{V}\) is selected for semantic scene modeling. Vehicles posing an interaction risk to the ego vehicle are defined as those whose predicted trajectories form potential conflict points with the ego trajectory at the intersection.

For each such conflict pair, the interaction risk is quantified using the Time-to-Collision Difference ($\Delta \mathrm{TTC}$). If the absolute value of $\Delta \mathrm{TTC}$ between the ego vehicle and another vehicle is smaller than a predefined threshold \(\Delta T_{th}\), the interaction is considered risky, and the corresponding vehicle is selected as a modeling object. The threshold is empirically determined based on common intersection safety practices, ensuring that vehicles with temporally close arrival at conflict points are retained for interaction modeling.

In addition to interaction risk, vehicles that are spatially proximate to the ego vehicle are also included. Spatial proximity is defined using the longitudinal distance along the lane centerline rather than Euclidean distance. Specifically, vehicles whose longitudinal distance to the ego vehicle is less than \(d_{th}\) are considered spatially proximate and are selected as modeling objects. This criterion allows the model to capture short-term influential vehicles while filtering out distant participants with negligible impact on the ego vehicle's decision-making.

Consequently, the selected modeling object set consists of vehicles that satisfy at least one of the following conditions:

(1) vehicles involved in potential trajectory conflicts with the ego vehicle and $\Delta \mathrm{TTC}$ is less than the threshold.

(2) vehicles located within a spatial proximity threshold of the ego vehicle.

\subsection{Elements and Formal Representation of Scene Modeling}

For traffic participants in general driving scenarios, their states can be formalized as a triple: \(\text{\{P,S,H\}}\) , where \(P = \{\left( x_{1},y_{1} \right),\left( x_{2},y_{2} \right)\ldots\left( x_{i},y_{i} \right)\text{\}}\)\emph{,} representing the positions of all vehicles within the scene at this moment, where \(S = \{\left( v_{1},a_{1} \right),\left( v_{2},a_{2} \right)\ldots\left( v_{i},a_{i} \right)\text{\}}\), represent the dynamic states of each vehicle in this scene at this moment, where \(H = \{ h_{1},h_{2}\ldots h_{i}\text{\}}\),represent the historical status of each vehicle.

Assuming recording range is \(T\), where \(h_{i} = \{\left( p_{i(t - T)},s_{i(t - T)}),(p_{i(t - T - 1)},s_{i(t - T - 1)})\ldots(p_{i(t)},s_{i(t)}) \right\}\), capturing the position and kinematic information of vehicle \(i\) from \(t - T\) to \(t\).

Considering the constraints of perception capabilities and computational resources, this study selects only target vehicles that pose an interaction risk or are spatially proximate to the ego vehicle as modeling objects, forming the object elements of the OPM framework.

For each object, its kinematic and spatial position states are sampled as follows:

\begin{equation}
\mathcal{F}_{\mathcal{i}} = \left\lbrack x_{i},y_{i},v_{i}^{x},v_{i}^{y},a_{i}^{x},a_{i}^{y},\phi_{i} \right\rbrack\tag{1}
\end{equation}

where \(x_{i},y_{i}\) represent the planar position coordinates of the vehicle, \(v_{i}^{x},v_{i}^{y},a_{i}^{x},a_{i}^{y}\ \)respectively represent the velocity and acceleration components in the coordinate system, \(\phi_{i}\) represents the vehicle's heading angle.

The current intersection morphology is defined as

\begin{equation}
\mathcal{I} =\text{\{}\mathcal{P}_{\text{stop}},\mathcal{R}_{lane},\mathcal{L}_{\mathbf{ref}}\text{\}}\tag{2}
\end{equation}

where \(P_{\text{stop}}\) represents the position of the intersection stop line, \(R_{\text{lane}} = \{ R_{1},R_{2},\ldots,R_{N}\text{\}}\) represents the boundaries of the intersection lanes, and\(\ \mathcal{L}_{ref}\) indicates the reference lines for each lane center.

Based on the vehicle's current position \(\left( x_{i},y_{i} \right)\) , the system identifies the lane \(R_{i}\) in which the vehicle is located. Subsequently, the absolute position is converted into a relative distance \(p_{i}\) from the intersection. The nearest point on the lane's central reference line is retrieved, and the direction vector t at that point is calculated. The angles between the vehicle's velocity and the lane direction, \(\theta_{v}\), and between the acceleration and the lane direction, \(\theta_{a}\), and then determined.

The current process state is ultimately assigned as follows:

\begin{equation}
P_{ti} = \left( p_{i},v_{i},a_{i},\theta_{vi},\theta_{ai},l_{i} \right)\ \ \tag{3}
\end{equation}

where \(p_{i}\) represents the vehicle's relative position to the intersection, \(v_{i}\) denotes the vehicle's speed, \(a_{i}\) indicates the vehicle's acceleration, \(\theta_{v}\) is the angle between the velocity and the lane direction, \(\theta_{a}\) is the angle between the acceleration and the lane direction, and \(l_{i}\) specifies the lane type and identifier.

\subsection{Interaction Relationship Modeling and Conflict Risk Assessment}

In an intersection, potential conflict relationships may exist among vehicles. To address this, the system traverses the set of process states \(P_{t}\) to detect whether any sub-state \(P_{ti}\) conflicts with the sub-states of other CAVs. This involves determining whether the current states of two vehicles could lead to a collision, thereby constructing an interaction pair \(C\left( v_{i},v_{j} \right)\).The risk level of the conflict is quantified using the Time-to-Collision Difference \(\Delta TTC\), a widely adopted metric for assessing the collision risk at potential conflict points between vehicles, defined as

\begin{equation}
{\Delta TTC}_{i,j} = \left| {TTC}_{i} - {TTC}_{j} \right| = \left| \frac{d_{i}}{v_{i}} - \frac{d_{j}}{v_{j}} \right|\tag{4}
\end{equation}

It should be noted that TTC-style indicators may become numerically unstable when the relative speed approaches zero. Therefore, in near-stop or creeping interaction states, the conflict estimate is interpreted with caution to avoid exaggerated risk values caused by near-zero denominators.

Additionally, it is assumed that vehicles within the scene can engage in direct intent-based interactions. When a vehicle makes decisions based on its intents, it must interact with other vehicles, resulting in the construction of intent-associated interaction pairs \(I\left( v_{i},v_{j} \right)\).

Ultimately, the current relationships among vehicles in the scene can be represented as

\begin{equation}
\mathcal{R}_{\mathcal{t}} = C\left( v_{i},v_{j},{\Delta TTC}_{i,j} \right),I\left( v_{i},v_{j} \right)\tag{5}
\end{equation}

\subsection{OPM Scene Graph Construction}

Based on the aforementioned derivations, a triplet describing the OPM framework can be obtained as

\begin{equation}
T = \left( \mathcal{O}_{\mathcal{t}},\mathcal{P}_{\mathcal{t}},\mathcal{R}_{\mathcal{t}} \right)\tag{6}
\end{equation}

These structured semantic units serve as input prompts for the LLM, replacing traditional low-level numerical features and enhancing the model's ability to comprehend complex multi-vehicle conflict scenarios. Compared with directly feeding low-level spatiotemporal features into the language model, the OPM representation explicitly organizes the scene into objects, processes, and relations before semantic reasoning is performed. In this way, part of the structural abstraction burden is shifted from the LLM itself to an interpretable semantic layer, reducing irrelevant feature interference and making the reasoning context more compact and causally meaningful. In addition, the effective importance of scene elements is jointly reflected by their interaction relevance, conflict significance, and state-dependent influence on the current decision scenario, which helps the LLM focus on semantically critical factors rather than raw feature volume.

\section{Integration of Decision-Making Process with LLMs}
\subsection{Intent Parsing of Conflicting Parties at Intersections}

Unsignalized intersections are typical scenarios with ambiguous right-of-way, high collision risks, and frequent decision-making failures for autonomous vehicles. This study selects the unprotected left-turn scenario as a representative verification scenario and introduces a LLM as a socialized intent parser to parse the intents of conflicting parties, thereby providing support for decision-making.

For the intent of the opposing party, we first assume that the information set at a specific moment in the scenario is \(I\), It is necessary to screen the surrounding vehicle information \(I_{neighbor}\) of the ego vehicle and the explicit information \(I_{explicit}\) related to the ego vehicle. Among them, \(I_{neighbor}\) is associated with the motion trend of surrounding vehicles: on the one hand, the motion trend directly reflects the vehicle's intent and exerts a direct impact on the behavior of interacting vehicles; on the other hand, it reflects the degree of influence of the vehicle on the current scenario. To measure this influence degree, by analogy with the attention mechanism used to evaluate the relationship between contexts, information such as vehicle acceleration, position variation, and relative distance is selected to construct a query vector \(Q \in (\nabla\mathcal{F}_{\mathcal{i}}\), \(\mathcal{F}_{\mathcal{i}})\). Subsequently, a key vector \(K\), for evaluating the influence degree of each piece of information is obtained through preliminary training. After importing \emph{Q} and \emph{K} into the attention mechanism formula, the criticality of the vehicle to the overall scenario can be calculated as

\begin{equation}
Attention(Q,K,V) = softmax\left( \frac{Q_{t}K_{t}^{T}}{\sqrt{d_{k}}} \right)V_{t}\tag{7}
\end{equation}

Specifically, the key vector is learned offline from structured interaction-related features so that the attention mechanism can evaluate the relative importance of surrounding vehicles to the current decision scenario. The resulting saliency scores are then combined with the ego vehicle's explicit intent and other agents' explicit intents, and these structured semantic signals are organized as inputs to the LLM for downstream reasoning and strategy recommendation.

\(I_{explicit}\) is associated with the explicit intents of vehicles. Explicit intents refer to information with clear meanings, including the vehicle's steering wheel data, turn signal information, and linguistic information directly received via the eHMI, among which linguistic information is the most critical. Such information, together with dynamic data, will be input into the LLM.

Unlike conventional models that rely solely on sensors to extract explicit variables, a LLM can fuse multiple information sources---such as dialogue history, linguistic style, and cultural semantic knowledge bases---to infer an opponent's ``unspoken intentions'' and the ``evolution of behavioral style.'' Examples include judging whether a driver is about to yield, whether they are exhibiting anxious or aggressive affect, and whether that style is likely to change in the short term.

Specifically, the LLM mines linguistic inputs to identify (i) participant style, (ii) the participant' s current actions, and (iii) the participant's task information. Style features include categories such as aggressive, balanced, and conservative, which determine how assertive the ego vehicle's next decision should be. Current actions encompass maneuver types such as steering and acceleration/deceleration, which directly shape the ego vehicle's subsequent actions. Task information reflects the vehicle's present objective---e.g., commuting, emergency affairs, or official duties---and serves as the basis for priority assessment, directly influencing whether the ego vehicle should adopt a passive or proactive stance in the scene.

This process can be regarded as the mapping of natural language to a three-dimensional vector, namely, 
\begin{equation}
I_{explicit} \rightarrow \left\lbrack I_{style},I_{action},I_{task} \right\rbrack\tag{8}
\end{equation}

Since explicit information does not fully cover all the aforementioned content, information gaps may exist. The degree of influence of such information (including both complete and incomplete parts) on the final decision-making can be expressed as: \(I_{task} > I_{action} > I_{style}\)\emph{.}

For the ego-vehicle's own intent, the LLM will proactively query the passenger, and the query results will also revolve around the aforementioned three dimensions. After acquiring information on these three dimensions, the ego-vehicle will obtain the explicit intent set \(\{ I_{explicitEgo},I_{explicitElse}\}\), which serves as the basis for subsequent decision-making.

\subsection{Initial Construction of the Decision Tree}

At intersections, vehicle decision-making is primarily influenced by the relative position of vehicles with respect to the intersection geometry rather than their absolute global coordinates. Under traffic rule constraints, the set of feasible maneuvers is determined by the current lane and the vehicle's position within that lane. In addition, task urgency and interaction context further shape the driver's decision tendency. Therefore, the decision-making process at intersections can be modeled as a mapping from the current scenario state and driver attributes to a discrete set of feasible actions. Specifically, in intersection scenarios, available decisions can be categorized into the following five types: \{\emph{straight-line acceleration, straight-line deceleration, straight-line constant speed, left turn, right turn}\}.

The decision-making process of a driver at an intersection can be viewed as a utility-based selection problem, in which alternative maneuvers are evaluated by jointly considering the driver's objectives (e.g., safety, efficiency, and comfort) and the level of interaction or conflict present in the current traffic scenario. From a behavioral perspective, the safety-related component reflects the tendency to avoid conflict and preserve maneuver stability under uncertainty, the efficiency-related component reflects the tendency to maintain passage progress and reduce delay, and the comfort-related component reflects the preference to avoid abrupt acceleration and deceleration. In this sense, the systematic utility is used not only as a mathematical scoring term, but also as a compact representation of the trade-offs underlying interactive driving behavior.

In transportation research and behavioral modeling, such decision-making is commonly formulated as a discrete choice process, where each candidate action is associated with a latent utility value, and the driver is assumed to select the option with the highest perceived utility. This framework originates from random utility theory, which has been widely used to model human decision-making under uncertainty, including driving behavior in interactive traffic environments (McFadden, 1974; Ben-Akiva and Lerman, 1985)

Formally, for a decision-maker \(k\) facing a choice set \(C\), the preference for option \(i\) over any other option \(j \in C\) can be expressed as

\begin{equation}
U\left( z_{i},s_{k} \right) \geq U\left( z_{j},s_{k} \right)\quad\forall j \in C\tag{9}
\end{equation}

where \(U\left( z_{i},s_{k} \right)\) denotes the utility of decision option \(i\); \(z_{i}\) and \(z_{j}\) represent the attribute vectors of options \(i\) and \(j\), respectively (e.g., maneuver type, expected safety level, and efficiency impact); and \(s_{k}\) characterizes the preference-related attributes of decision-maker \(k\) such as task urgency or safety awareness.

Under this formulation, the driver's behavior is not assumed to be perfectly deterministic. Instead, decision outcomes are modeled probabilistically to capture unobserved factors and bounded rationality, which is consistent with empirical findings in driver behavior modeling and discrete choice theory.

For decision-making at intersections, we evaluate each candidate maneuver using three criteria: matching degree, safety degree, and efficiency degree. The matching degree reflects how well a maneuver conforms to the current intersection context (e.g., presence of a leading vehicle) and the route guidance at the ego vehicle's current location. We assign equal baseline weights to the five maneuver candidates (in the predefined order), meaning that all options start with the same prior preference before applying scenario-dependent penalties or adjustments.

When a leading vehicle is detected, we compute a car-following penalty using the Intelligent Driver Model (IDM) and accordingly down-/up-weight the straight-line maneuver candidates to reflect feasibility and interaction constraints imposed by the leader. Subsequently, we check whether a candidate maneuver is consistent with the navigation instruction (e.g., required turning direction at the current approach). If consistent, the original weight is retained; otherwise, the candidate receives a substantial penalty to discourage route-infeasible actions.

The safety degree represents the relative emphasis on safety in the current trip and is modeled as a preference-related factor that depends on both the ego vehicle's explicit intent \(I_{explicitEgo}\) (e.g., urgency) and the scenario context (e.g., traffic density and conflict intensity). Intuitively, higher safety preference leads to more conservative maneuver choices, consistent with established behavioral modeling practice where preferences modulate the utility of alternatives rather than deterministically selecting a single action. The efficiency degree captures the preference for timely passage, influenced by the ego occupants' urgency and, when applicable, the urgency or priority cues of interacting road users represented by \(I_{explicitElse}\). In conflict situations, a higher efficiency preference increases the utility of maneuvers that resolve conflicts promptly and reduce prolonged mutual interference.

Based on these criteria, we define the systematic (deterministic) utility for maneuver \(\mathbf{i}\) as a linear-in-parameters function:

\begin{equation}
V_{i} = \beta^{T}x_{i}\tag{9}
\end{equation}

where \(x_{i}\) aggregates the maneuver-specific attributes (including matching-, safety-, and efficiency-related features as well as interaction indicators), and \(\beta\) is the corresponding coefficient vector.

To account for unobserved factors (e.g., latent driver style variations, perception noise, and bounded rationality), we adopt the random utility formulation:

\begin{equation}
U_{i} = V_{i} + \epsilon_{i}\tag{10}
\end{equation}

where \(\epsilon_{i}\) \hspace{0pt} is an i.i.d. stochastic term. Under the standard assumption that \(\epsilon_{i}\) follows a type-I extreme value (Gumbel) distribution, the resulting discrete choice model reduces to the multinomial logit model.

Therefore, the baseline probability of selecting maneuver \(i\) from the choice set is:

\begin{equation}
P_{ni} = \frac{e^{V_{ni}}}{\sum_{j}^{}e^{V_{nj}}}\tag{11}
\end{equation}

The probabilities derived from Eq. (11) serve as the baseline maneuver-selection distribution, which is subsequently refined by the semantic reasoning and correction modules described later.

\subsection{Decision Tree Correction}

Although the logit-based decision probabilities derived from the instantaneous system utility can reasonably approximate the optimal choice under the current traffic state, empirical studies have consistently shown that human driving behavior is not purely myopic. Instead, drivers tend to exhibit temporal consistency and path continuity, maintaining short-term plans and avoiding frequent or abrupt changes in maneuver selection, even when the instantaneous utility differences are small. This sequential and plan-driven nature of driving behavior has been widely acknowledged in microscopic traffic and behavioral modeling literature. Therefore, incorporating historical information into the decision-making process is essential to obtain behaviorally realistic and stable maneuver choices.

Let \(\mathcal{H = \{}h_{1},h_{2}\ldots h_{i}\text{\}}\), denote the set of historical states of the ego vehicle within a recent time window. To model behavioral inertia and short-term planning effects (Toledo et al., 2007), we introduce a historical consistency term that adjusts the current systematic utility:

\begin{equation}
\widetilde{V_{ni}} = V_{ni} + \gamma \cdot D_{ni}^{\text{prev}}\tag{12}
\end{equation}

where \(\gamma\) is a historical consistency coefficient controlling the influence of past decisions on the current choice. \(D_{ni}^{\text{prev}}\)\hspace{0pt} represents the tendency to repeat maneuver \(i\), which may be interpreted as the persistence of the driver's short-term plan or action preference.

Specifically, \(D_{ni}^{\text{prev}}\) is computed by measuring the similarity between the current scenario state and past states in which maneuver \(i\) was selected. Following common practice in sequential decision modeling, we define this tendency as the maximum cosine similarity

\begin{equation}
D_{ni}^{\text{prev}} = \max_{t \in \mathcal{H}_{\mathcal{i}}}\frac{\mathbf{x}_{\mathbf{n}} \cdot \mathbf{x}_{\mathbf{n}}^{\left( \mathbf{t} \right)}}{\text{|}\mathbf{x}_{\mathbf{n}}\text{|} \cdot \text{|}\mathbf{x}_{\mathbf{n}}^{\left( \mathbf{t} \right)}\text{|}}\tag{13}
\end{equation}

where \(\mathcal{H}_{\mathcal{i}}\) denotes the set of historical time steps at which maneuver \(i\) was executed, \(\mathbf{x}_{\mathbf{n}}\) is the feature vector of the current scenario state, and \(\mathbf{x}_{\mathbf{n}}^{\left( \mathbf{t} \right)}\)\hspace{0pt} is the corresponding state feature at time \(\mathbf{t}\). This formulation encourages maneuver choices that are consistent with previously observed driving contexts, thereby enhancing temporal coherence.

In addition to historical preference, drivers also tend to avoid unnecessary maneuver switching, as frequent changes can increase cognitive load and perceived risk. To reflect this behavioral tendency and suppress oscillatory decisions caused by marginal utility fluctuations, we introduce an action-switch penalty:

\begin{equation}
\widetilde{V_{ni}} = V_{ni} - \lambda \cdot \mathbb{I}(i \neq i^{\text{prev}})\tag{14}
\end{equation}

where \(i^{\text{prev}}\ \)is the maneuver selected at the previous time step; \(\mathbb{I}(\bullet)\) is an indicator function, and \(\lambda\) is a smoothing coefficient penalizing unnecessary action changes. Similar continuity-enforcing mechanisms have been adopted in sequential driving and risk-aware behavior models to ensure realistic maneuver evolution over time (Kesting et al., 2007).

After incorporating historical consistency and action smoothness, the corrected maneuver-selection probability is obtained via the logit formulation:

\begin{equation}
P_{ni} = \frac{e^{\widetilde{V_{ni}}}}{\sum_{j}^{}e^{\widetilde{V_{nj}}}}\tag{15}
\end{equation}

However, traffic scenarios often involve implicit, unquantifiable, and socially conditioned intent information that cannot be fully captured by structured state variables or hand-crafted utility terms. To address this limitation, we further embed a Large Language Model (LLM) into the behavior decision-making process as a semantic reasoning and correction module. The LLM receives multi-source inputs, including the saliency vector \(Attention(Q,K,V)\), the ego vehicle's explicit intent \(I_{explicitEgo}\) , and other agents' explicit intents \(I_{explicitElse}\), and performs high-level semantic parsing and intent-conflict reasoning.

Through prompt-based natural language reasoning, the LLM evaluates intent priority, behavioral feasibility, and social rationality, and outputs an optimal strategy recommendation \(\mathcal{A}_{\mathbf{final}}\) , which is used to refine and correct the initial behavior tree. This integration enables the modeling of implicit, non-linear, and context-dependent factors that are difficult to formalize explicitly, thereby improving adaptability and human-likeness in complex interactive scenarios . The overall correction process can be expressed as:

\begin{equation}
\left( \mathcal{A}_{\mathbf{final}} \right)\mathbf{=}LLM\left( P_{n}\mathbf{,}Scene,Prompt \right)\tag{16}
\end{equation}
 typical Prompt example is shown below:

\begin{table*}[!t]
\caption{LLM enhanced decision optimization prompt Template}
\label{tab:prompt_template}
\centering
\small
\begin{tabular}{p{0.97\textwidth}}
\toprule
\textbf{Parameters:} Selection probability $P_n$, scenario description $Scene$.\\
\textbf{Output:} Optimal strategy recommendation $\mathcal{A}_{\mathrm{final}}$.\\
\textbf{[Role Setting]} You are an experienced driver currently in an unsignalized intersection scenario. Below is a summary of information output by your current perception and decision-making modules.\\
\textbf{[Input information summary]}\\
1. Scene structured perception information (Scene): The oncoming vehicle is making a slow left turn, with a current speed of 3.2 m/s and an angle of 30$^{\circ}$ between its acceleration direction and the lane line. Based on the calculation result of the attention mechanism, the importance score of this vehicle to the current decision-making scenario is 0.83, which is derived from the computation of $Attention(Q,K,V)$.\\
2. Candidate Behavior Options and Utility Scores: Your five candidate behaviors and their corresponding Logit model scores (utility values $P_{ni}$) are as follows: straight-line acceleration (0.32), straight-line deceleration (0.48), constant speed maintenance (0.10), left turn (0.05), right turn (0.05).\\
Contextual Intent Input: The passenger has just stated: ``I am in a hurry and hope to pass through as soon as possible.''\\
\textbf{[Task objective]} Judgment results based on the provided information: (i) optimal current behavior selection and its rationale; (ii) candidate behaviors with potential conflicts or needing pruning.\\
\bottomrule
\end{tabular}
\end{table*}

LLMs may generate decision recommendations that seem reasonable but are inconsistent with facts based on incomplete, ambiguous, or even irrelevant inputs---a phenomenon known as "hallucination." In autonomous driving, this could manifest as incorrect inference of non-existent intents. For example, normal lane-changing behavior might be mistakenly judged as an emergency obstacle-avoidance intent, leading to erroneous operations. To address this, the following measures are adopted:

(1) Introduction of an Intent Recognition Confidence Threshold:A confidence metric is introduced. When the confidence of the LLM's intent parsing is lower than a specific threshold (e.g., 0.7), the current intent is labeled as a "weak intent" and prohibited from being used in the main decision-making logic. Here, the threshold of 0.7 is used as a practical balance between semantic sensitivity and decision reliability. A lower threshold would allow more uncertain intent interpretations to enter the decision-making process and may increase unstable semantic corrections in ambiguous scenarios, whereas a higher threshold would suppress too many potentially useful interaction cues and make the decision layer overly conservative. Therefore, the current threshold is adopted as a moderate operating point for filtering weak semantic inferences in mixed-traffic interactions.

(2) Introduction of a Human Feedback Loop:A "confirmatory interaction" is integrated when simulating passenger intent input. For instance, the system may ask, "Do you want me to accelerate to pass through?" This enhances the intent verification process and reduces risks caused by ambiguity. Nevertheless, in extreme conflict situations, different surrounding-driver styles may still lead to different semantic mismatch patterns. For aggressive surrounding drivers, the system may insufficiently capture abrupt insertion or forcing tendencies; for conservative surrounding drivers, the decision layer may become overly yielding and reduce traffic efficiency. Therefore, the current confidence-threshold mechanism and confirmatory interaction should be understood as ambiguity-reduction mechanisms rather than complete safeguards against all style-dependent failure cases.

The baseline model in this study is the \emph{Llama3-7B} large language model. The integration of the LLM effectively compensates for the limitations of traditional rule-driven methods in handling semantic consistency, behavioral coherence, and unstructured inputs. Leveraging its ability to understand natural language and high-dimensional semantics, the LLM endows the system with context-construction capabilities and strategy-maintenance tendencies similar to human drivers. It can reasonably infer the potential logical relationships between interaction history, current intents, and future behaviors in continuous dynamic scenarios.

\subsection{Trajectory Generation and Optimization}

After constructing and pruning the behavior tree via the OPM (assumed as the aforementioned decision optimization framework) to obtain the final decision, safety constraints, and efficiency constraints, it is necessary to generate an optimal trajectory based on the decision. To this end, we propose a Monte Carlo trajectory generation optimization method enhanced by a LLM.

A fifth-order polynomial is used to describe the vehicle's behavior. Based on the lane reference line, the vehicle's path distance along the lane is defined as s, and the lateral distance relative to the reference line is defined as l. These two variables are expressed as functions of time t, i.e.,

\begin{equation}
s(t) = a_{0} + a_{1}t + a_{2}t^{2} + a_{3}t^{3} + a_{4}t^{4},\tag{17}
\end{equation}

\begin{equation}
l(t) = b_{0} + b_{1}t + b_{2}t^{2} + b_{3}t^{3} + b_{4}t^{4} + b_{5}t^{5}\tag{18}
\end{equation}

The corresponding velocity, acceleration, and other dynamic parameters can be expressed as follows:

\begin{equation}
s\left( t_{0} \right) = s_{0},\quad\dot{s}\left( t_{0} \right) = v_{s0},\quad\ddot{s}\left( t_{0} \right) = a_{s0},\tag{19}
\end{equation}

\begin{equation}
\dot{s}(T) = v_{sT},\quad\ddot{s}(T) = a_{sT}\tag{20}
\end{equation}

\begin{equation}
l\left( t_{0} \right) = l_{0},\quad\dot{l}\left( t_{0} \right) = v_{l0},\quad\ddot{l}\left( t_{0} \right) = a_{l0},\tag{21}
\end{equation}

\begin{equation}
l(T) = l_{T},\quad\dot{l}(T) = v_{lT},\quad\ddot{l}(T) = a_{lT}.\tag{22}
\end{equation}

In complex mixed driving environments, the interaction relationships between traffic participants are intricate, making it difficult to directly calculate the optimal trajectory. To address this, we reformulate trajectory search as a sampling-based optimization problem; Monte Carlo sampling generates candidate terminal states, and a scoring function under safety/efficiency constraints selects the best trajectory. Based on the principle that vehicle speeds in traffic scenarios follow a Gaussian distribution, we predict a set of future states of the ego-vehicle using Monte Carlo sampling (based on the standard Gaussian distribution) and combined with the ego-vehicle's initial state. Mathematically, this is expressed as

\begin{equation}
s_{end} = f_{a}\left( s_{start} \right) \odot \left( 1 + \mathcal{N}(0,1) \right)\tag{23}
\end{equation}

Then, calculate the corresponding trajectory:

\begin{equation}
\tau_{k} = QuinticPoly\left( s_{start},s_{end} \right)\tag{24}
\end{equation}

Finally, the duration, average speed, and average acceleration indicators corresponding to this trajectory will be calculated. The weighted loss function combining safety constraints and efficiency constraints will be computed, and the optimal trajectory will be selected. From a computational perspective, the cost of this sampling-based optimization increases with the number of sampled terminal states, candidate actions, and trajectory evaluations under the safety and efficiency constraints. As the interaction scale grows, the repeated generation and scoring of candidate trajectories correspondingly increase the runtime burden of the optimization layer.

The reasoning process described above is presented in Table 2:

\begin{table*}[!t]
\caption{LLM enhanced decision optimization for hazard-zone avoidance algorithm}
\label{tab:mc_opt}
\centering
\small
\begin{tabular}{p{0.05\textwidth} p{0.9\textwidth}}
\toprule
\textbf{Step} & \textbf{Description}\\
\midrule
1 & Initialize $\tau^* \leftarrow \varnothing$, $S_{\min} \leftarrow \infty$.\\
2 & Monte Carlo sampling loop.\\
3 & For $k = 0$ to $N_{\mathrm{samples}}$ do:\\
4 & Action selection: $a_k \sim A(s_{start}, R)$.\\
5 & Disturbed final state: $s_{end} \leftarrow f_a(s_{start}) \odot (1 + \mathcal{N}(0,1))$.\\
6 & Generate candidate trajectory: $\tau_k \leftarrow QuinticPoly(s_{start}, s_{end})$.\\
7 & Compute indicators: $T_{avg} = \frac{1}{n}\sum_{t=0}^{t_{end}}\Delta t$, $V_{avg} = \frac{d}{T_{avg}}$, $j_{avg} = \frac{1}{n}\sum_{t=0}^{t_{end}}\Delta t\cdot\left|a_{t+1} - a_t\right|$.\\
8 & Normalization and scoring: $S_k = c_e\frac{T_n}{T_{ref}} + c_s\frac{\int_0^{T_n} |\mathrm{jerk}(t)|\,dt}{J_{\max}}$.\\
9 & If $S_k < S_{\min}$ then:\\
10 & Update $\tau^* \leftarrow \tau_k$, $S_{\min} \leftarrow S_k$.\\
11 & End if.\\
12 & End for.\\
\bottomrule
\end{tabular}
\end{table*}

\section{Language-based Interaction via LLM and eHMI}
The basic process is as follows:

(1) Scenario Perception and Intent Parsing: The system receives structured scenario information and intent prompts from traffic participants, completing behavior reasoning and optimal strategy generation.

(2) Decision Language Translation: The decision results are input into the LLM again, converting the machine decision logic into natural language descriptions that conform to human cognitive habits. For instance, the machine decision of "straight-line deceleration" can be translated into natural language as: "I am slowing down; please go ahead."

(3) Language Interaction Loop: The system can conduct continuous multi-turn language interactions based on external feedback. The LLM will continuously receive the other party's information and update it to achieve intent clarification, emotion adaptation, and strategy adjustment. For instance, after identifying the emotional state of pedestrians or occupants (such as anxiety or hesitation), the system can adjust the tone and wording; when information is uncertain, the vehicle will proactively initiate inquiries, such as: "Do you intend to pass first?"

(4) Interactive Visual Output: The above interaction process will be presented externally in text form through the eHMI, and will also realize audio-visual interaction. Finally, this solution can achieve a closed loop from language to behavior, then to language, and back to behavior.

(5) Continuous Learning: After completing the dialogue, the LLM will continuously fine-tune the language model using interaction data, making it more in line with the context and habits of the target usage area.

By introducing the language-layer interaction process supported by LLMs, the interactive autonomous driving system constructed in this study breaks through the closed nature of the traditional perception-decision-control chain. It realizes active conversational collaboration for human traffic participants, providing strong support for the safe operation of autonomous driving in mixed driving environments.

\section{Experimental Results and Analysis}

To verify the autonomous driving interactive decision-making model in complex mixed driving environments, we conducted experiments using the Tongji University Cluster Driving Simulator (as shown in Fig. 3), which is consistent with commonly used public autonomous driving testing platforms (Sun et al., 2024). The model's performance was comprehensively evaluated from multiple dimensions, including safety, efficiency, and comfort. Finally, a Turing test was carried out to assess the model's human-like degree.

\begin{figure}[!t]
\centering
\includegraphics[width=\linewidth]{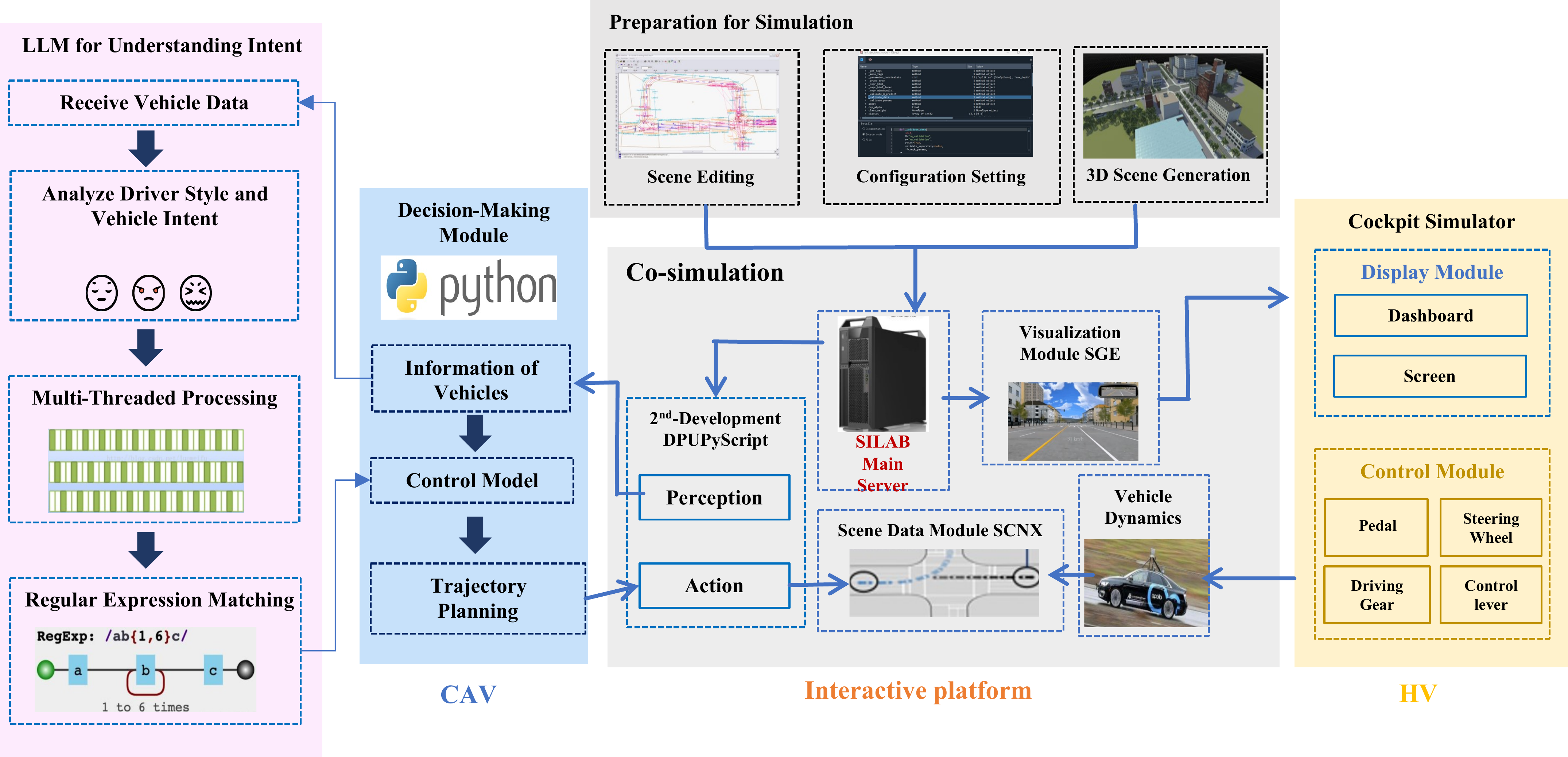}
\caption{Driving Simulator System}
\end{figure}

\subsection{Comprehensive Performance Analysis of the Autonomous Driving Decision Model}
\subsubsection{Experimental Scheme}

According to the 2022 Beijing Autonomous Driving Road Test Report, 51\% of autonomous driving takeovers occur at intersections. Therefore, this study constructed an intersection scenario in the SILAB driving simulator. To evaluate the comprehensive performance of the proposed LLM-based interactive autonomous driving decision method, the following experimental scheme was designed and implemented:

(1) Experimental Setup: The intersection center was set as the origin, with the centerlines of the two perpendicular roads serving as the horizontal and vertical axes, respectively. The initial coordinates of the HDV and CAV were (-40m, -5m) and (40m, 5m), respectively. The HDV traveled straight from left to right, with its trajectory controlled in real time by the same human driver operating the simulator to enable interaction between the HDV and CAV. The CAV made a left turn at the intersection, controlled by three different autonomous driving decision models: IDM car-following model (based on inter-vehicle distance); game-theoretic (GT) model; The proposed model (based on LLM). Three initial speeds were set---low speed (6 m/s), medium speed (8 m/s), and high speed (10 m/s)---with an initial acceleration of 0 m/s$^2$ for all cases.

(2) Experimental Process:

The HDV and CAV settings were initialized in the SILAB driving simulator software.

The HDV was controlled by the same human driver, while the CAV was controlled by each of the three models separately.

Cross-experimental verification was conducted under the three initial speed modes, with each experiment repeated 30 times.

(3) Evaluation Indicators:

The CAV's comprehensive performance at the intersection was evaluated from three dimensions: traffic efficiency, riding comfort, and traffic safety. The specific indicators are as follows:

Average Speed: This metric is used to measure the CAV\textquotesingle s traffic efficiency and is defined as the expected value of the CAV\textquotesingle s speed when traversing the intersection.

Average Jerk: Reflects riding comfort, defined as the mathematical expectation of the derivative of acceleration. A smaller absolute value indicates better comfort.

Average Conflict Duration: Used to evaluate the model's conflict resolution performance in potential collision risk scenarios, defined as the average duration from the emergence of a conflict point between the two vehicles to the resolution of the conflict point.

\subsubsection{Statistical analysis of the experimental results}

Fig. 4 presents the statistical analysis of the key performance indicators of the three decision-making models under different initial speed conditions. The specific analysis is as follows:

\begin{figure*}[!t]
\centering
\subfloat[]{\includegraphics[width=0.32\textwidth]{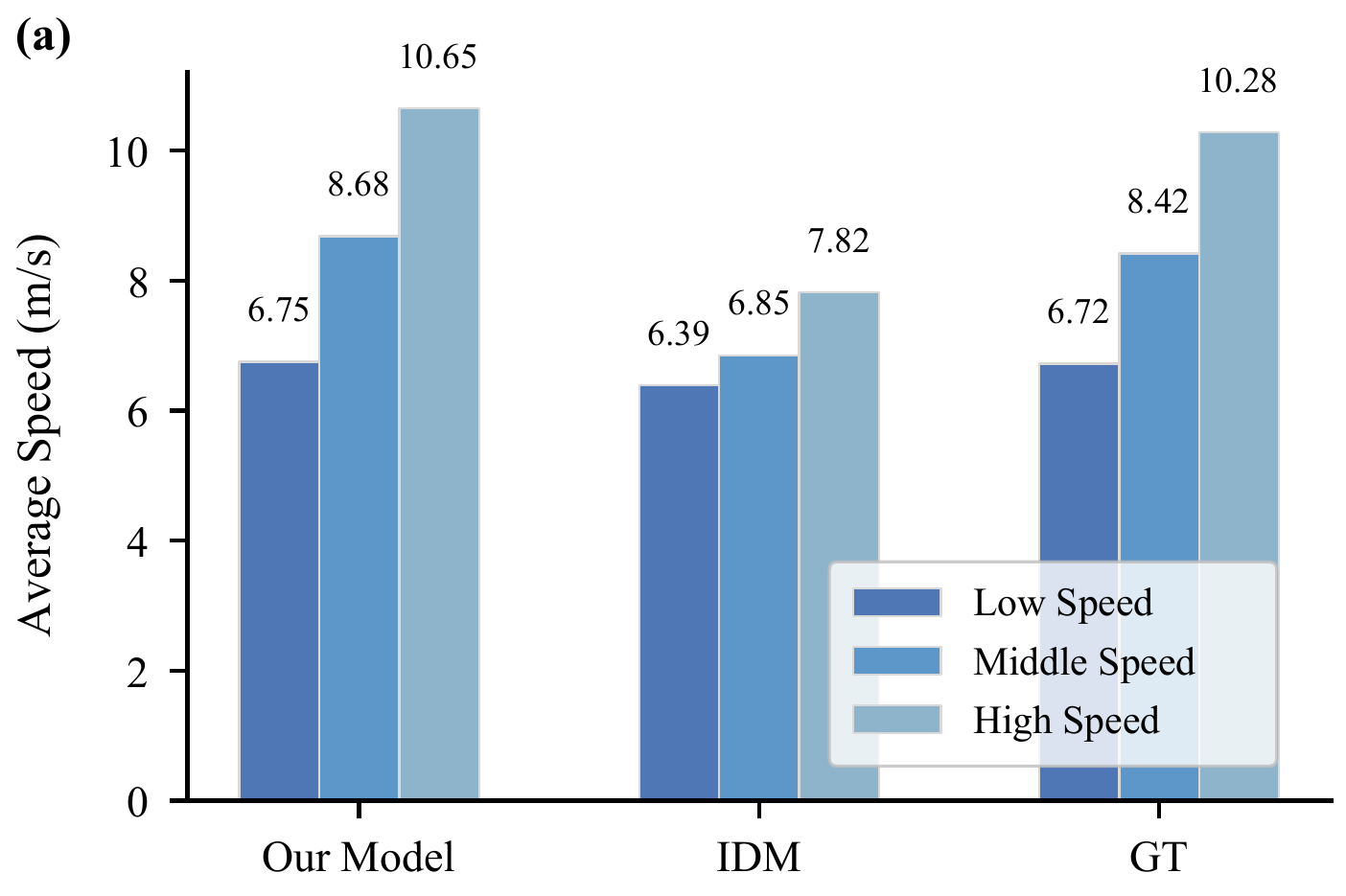}}
\subfloat[]{\includegraphics[width=0.32\textwidth]{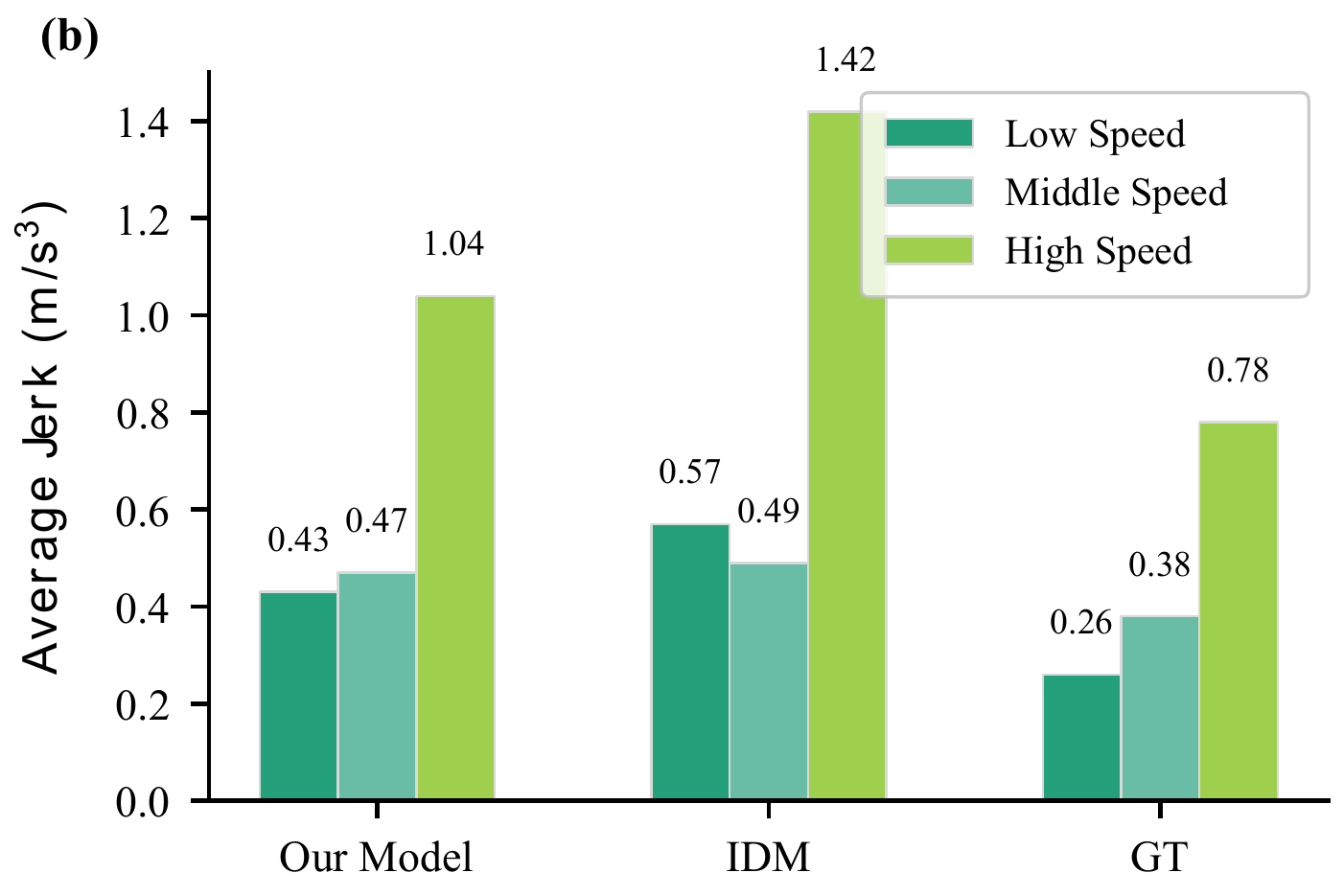}}
\subfloat[]{\includegraphics[width=0.32\textwidth]{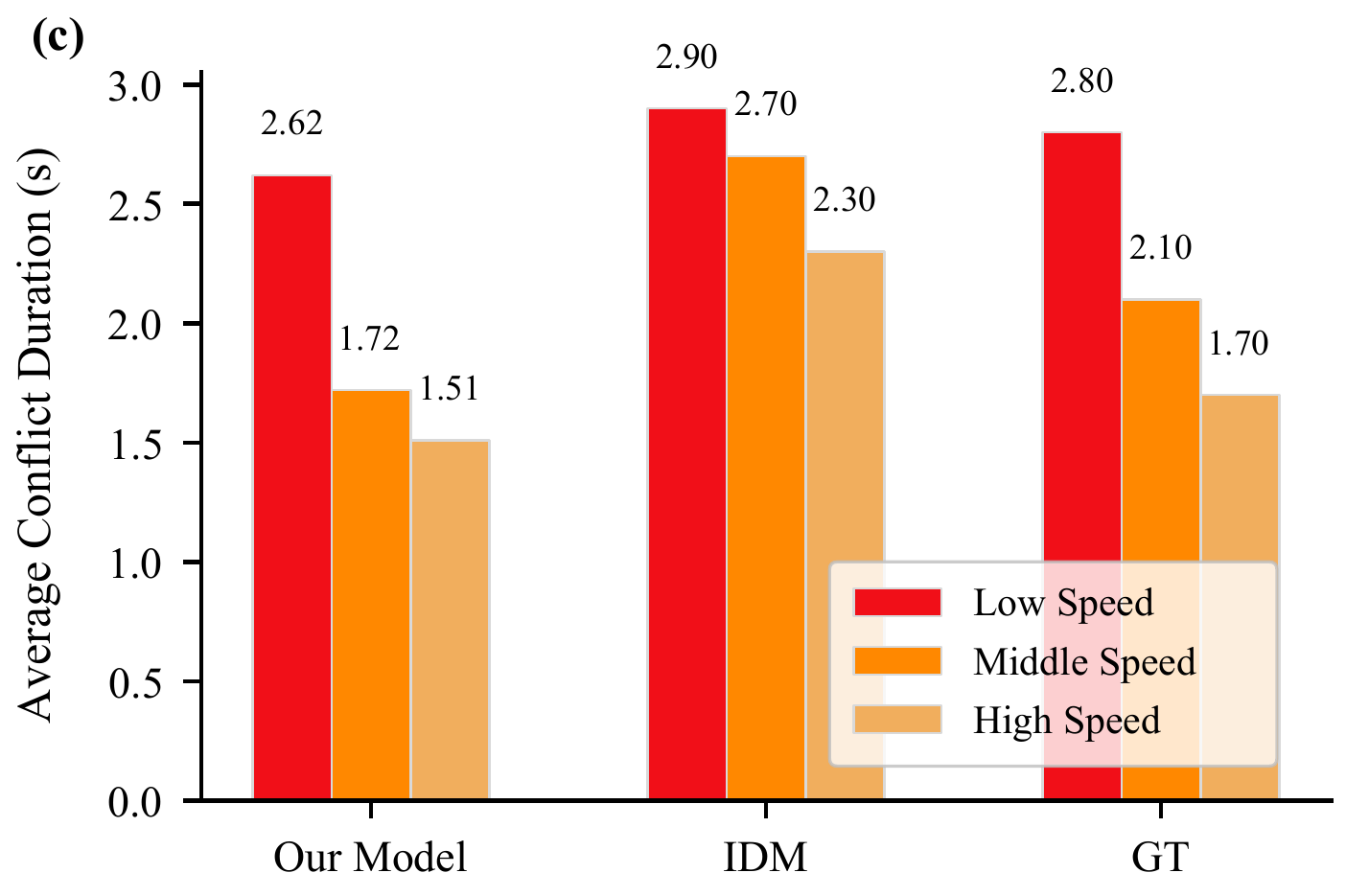}}
\caption{Comparison of three autonomous driving decision models under different initial speed conditions. (a) Average speed across models. (b) Average jerk across models. (c) Average conflict duration across models}
\label{fig:fig4}
\end{figure*}
The results of the driving efficiency statistics analysis are shown in Fig. 4(a). The performance comparisons of the three models under different initial speed modes are consistent. The average speed of the IDM car-following model is always at the lowest level, demonstrating a conservative decision-making characteristic. In contrast, the game-theoretic model based on game theory has a higher average speed, reflecting higher traffic efficiency. The proposed model achieves a higher average speed than the other two approaches across all three initial speed conditions, and has a more significant advantage in the driving efficiency indicators.

The results of the comfort statistics analysis are shown in Fig. 4(b). Under the three different initial speed conditions, the average jerk of the IDM car-following model is the highest. Especially in the high-speed scenarios, it fluctuates greatly, indicating that it is prone to sudden acceleration and deceleration behaviors in dynamic environments. In contrast, the game-theoretic model shows a lower acceleration value, attributed to the design of the comfort weight in the benefit function and the constraints on control variables. The proposed model has a slightly higher acceleration than the game-theoretic model, but is significantly superior to the IDM car-following model in terms of comfort, which is within the acceptable range.

The results of the safety statistics analysis are shown in Fig. 4(c). From the average conflict duration indicator, the model proposed in this paper has a significant advantage under all three initial speed conditions. Specifically, in the medium and high-speed scenarios, it shortens the average conflict duration by approximately 35\% compared to the IDM car-following model and by approximately 10\% compared to the game-theoretic model. This indicates that in the conflict decision-making at intersections, the proposed model not only maintains efficient traffic flow but also has the best safety performance, reduces overall traffic delay, and effectively alleviates potential congestion caused by conflicts.

\subsubsection{Comparative Analysis of OPM}

To quantitatively assess whether the proposed OPM semantic-structured representation yields tangible benefits beyond simple reformatting, a comparative study was conducted by varying only the LLM input format while keeping the evaluation scenarios and model configuration identical to the rest of the experimental settings in this paper. Three input formats were considered: OPM, the proposed object--process--relation representation, Raw, a baseline that directly feeds unstructured numeric or perceptual outputs, and Simple, a lightweight field-based organization that does not explicitly model object--process--relation semantics. Performance was evaluated using Accuracy, end-to-end inference Latency, and Reasoning Steps.

The results are summarized as Fig. 5 shows:

\begin{figure*}[!t]
\centering
\includegraphics[width=\textwidth]{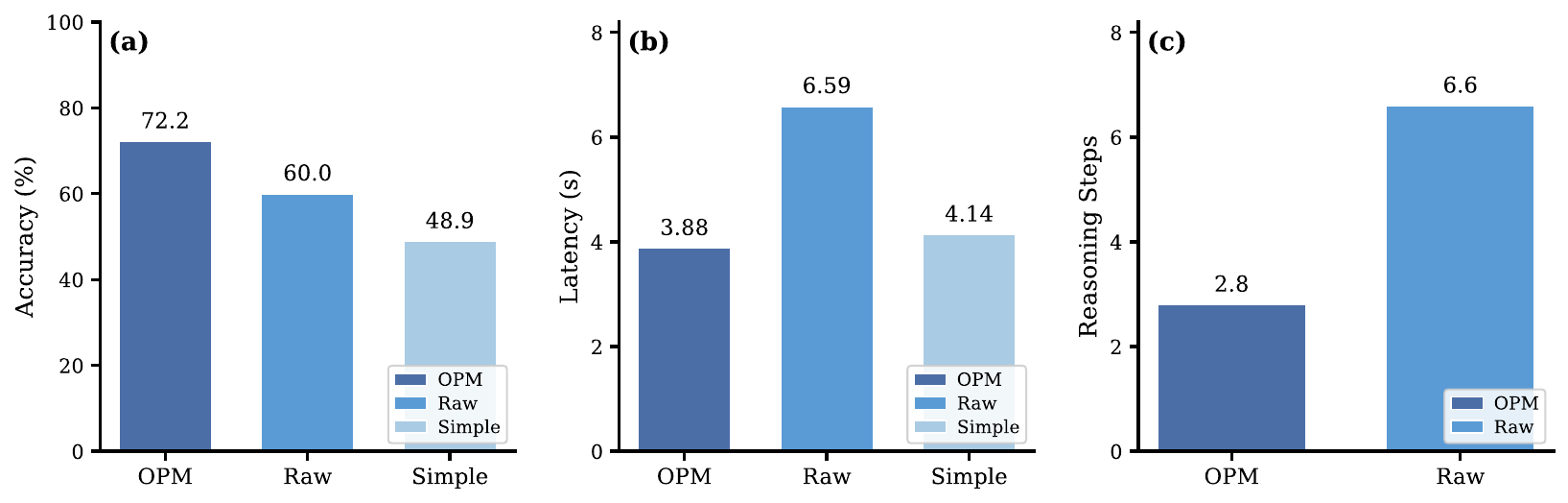}
\caption{Comparison of OPM with other methods. (a) Comparison of accuracy. (b) Comparison of latency. (c) Comparison of Reasoning steps.}
\label{fig:fig5}
\end{figure*}

The findings show that OPM improves accuracy by 12.2 percentage points over Raw while simultaneously reducing average latency from 6.59 s to 3.88 s, indicating that semantically structured inputs can reduce irrelevant information interference and enable more efficient inference under the same model setting. This result also suggests that a substantial portion of the decision latency is associated with the semantic reasoning burden imposed on the LLM. By organizing scene information into object--process--relation units, the proposed OPM representation helps shorten the effective reasoning path and reduce unnecessary prompt complexity, but the current inference time is still relatively high for a real-time interactive driving system. Therefore, the present framework should be understood primarily as a proof-of-concept validation of LLM-enabled interactive decision-making, rather than a fully optimized low-latency onboard deployment solution. In addition, OPM consistently outperforms Simple in accuracy, suggesting that superficial field organization alone is insufficient; explicitly representing object--process--relation semantics is more effective for capturing critical interaction factors and spatiotemporal constraints relevant to decision-making. Notably, for Simple, the model outputs did not consistently provide a stable, machine-countable reasoning-step field, and thus Reasoning Steps is reported as N/A.

Overall, this comparative experiment demonstrates that OPM provides measurable gains in both decision performance and inference efficiency relative to raw and shallowly structured alternatives, supporting its practical value as an input semantic structuring method.

\subsubsection{Case Analysis of Typical Examples}

To illustrate the performance differences, we present a comparative case analysis of the three autonomous driving decision models at an initial speed of 6 m/s. Fig. 6 shows the decision results of the three models at multiple moments in this scene.

\begin{figure*}[!t]
\centering
\includegraphics[width=\textwidth]{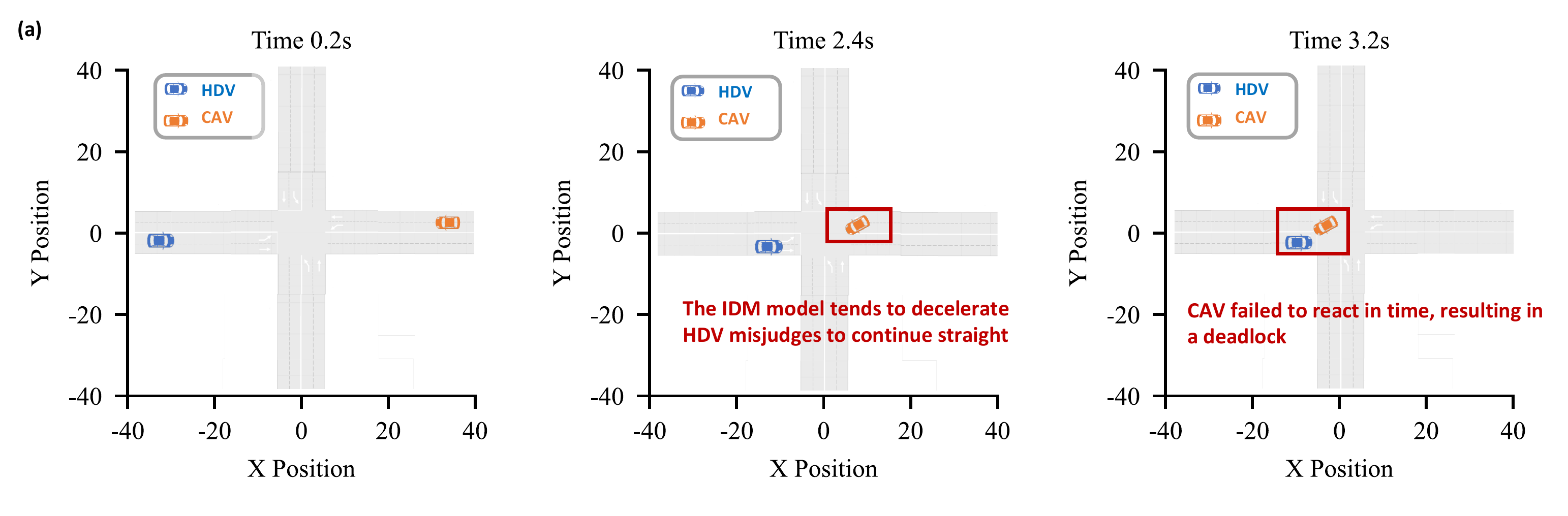}\\
\includegraphics[width=\textwidth]{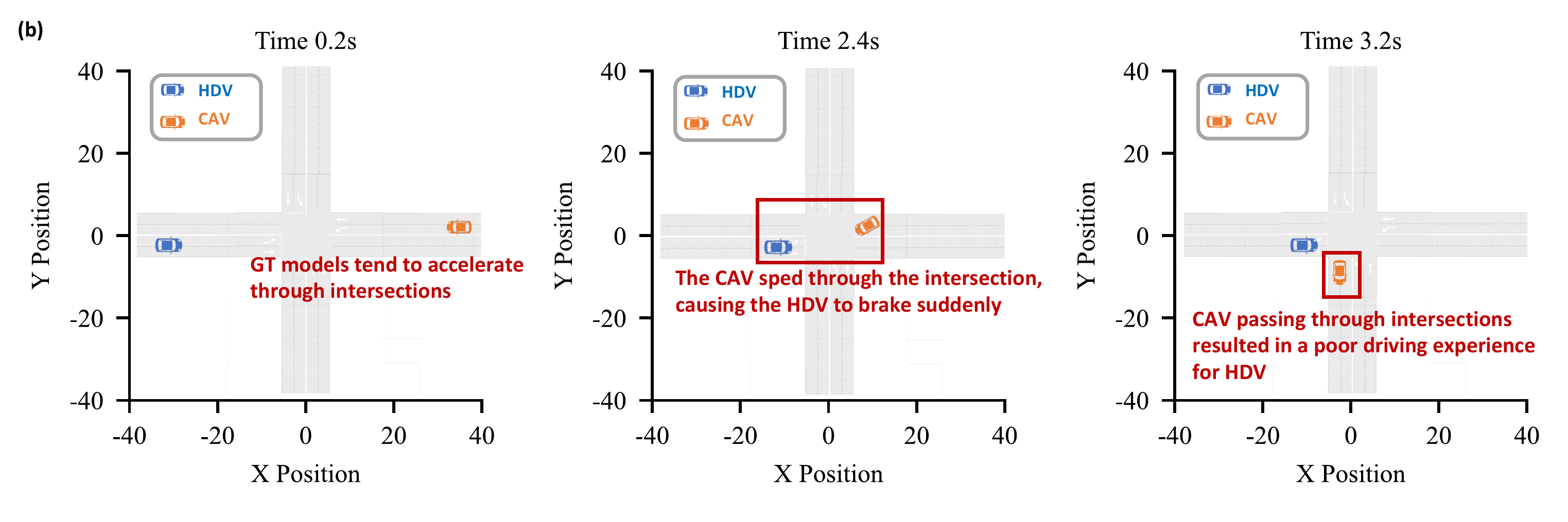}\\
\includegraphics[width=\textwidth]{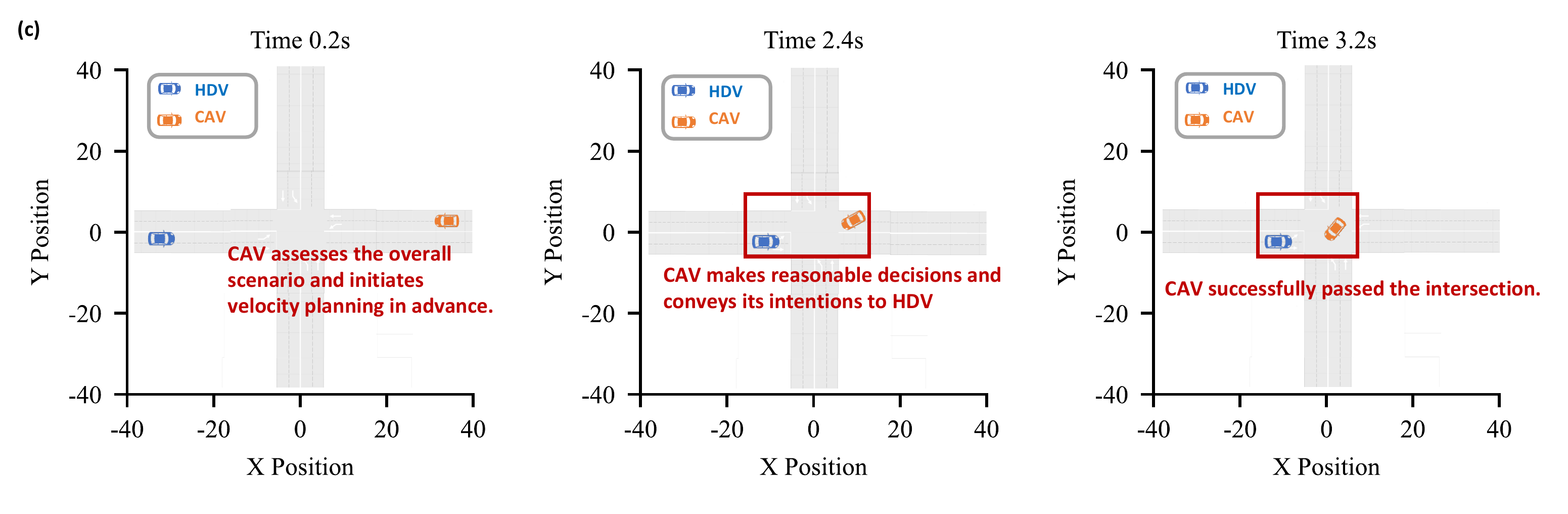}
\caption{Comparison of decision-making models for CAV at an initial velocity of 6 m/s.(a) Real-time vehicle positions under IDM. (b) Real-time vehicle positions under GT. (c) Real-time vehicle positions under LLM.}
\label{fig:fig6}
\end{figure*}

Fig. 6(a) shows the entire process of interaction between CAV and HDV under the control of the IDM car-following model. Due to the passive response of the IDM car-following model to the movement results of HDV using lane rotation projection, the decision-making is overly conservative. Therefore, a deceleration decision was adopted, resulting in a deadlock at the intersection and a driving conflict, which reduced the traffic efficiency.

Fig. 6(b) illustrates the entire process of interaction between CAV and HDV under the control of the game-theoretic model. Due to the tendency of this game-theoretic model to make aggressive decision-making through game theory, it accelerates the passage of the intersection, making it difficult for HDV to respond in time and forcing it to brake urgently. This leads to a higher safety risk and reduces the interaction experience of HDV.

Fig. 6(c) shows the entire process of interaction between CAV and HDV under the control of the LLM model. Since LLM continuously understands and analyzes the conflict scenarios and shares the decision-making intents with the human driver of HDV through eHMI, it achieves safe decision-making without sacrificing driving efficiency to a great extent through intent negotiation based on LLM.

In addition to unprotected intersections, which are widely regarded as the most representative scenarios for evaluating interactive decision-making capability, we further consider typical traffic scenarios to examine the generalization of the proposed method.

Among various complex driving scenarios, merging and roundabout scenarios share similar interaction characteristics with unprotected intersections, as vehicles must negotiate temporal access to shared space under limited structural constraints. However, compared with intersections, these scenarios introduce distinct features, such as stronger lane geometry constraints, asymmetric priority relationships, and localized conflict regions, which require the decision-making system to balance efficiency and safety through implicit intent coordination rather than explicit right-of-way rules. In such strongly coupled interactions, an error in the intent interpretation of one participant may influence the semantic understanding of the overall scenario and thereby affect downstream maneuver selection. This issue is not fully resolved by the current simulator validation and remains an important limitation when considering broader scenario generalization.

In contrast, congestion scenarios are dominated by longitudinal car-following dynamics, as also observed in congestion-oriented lane-changing and interaction studies (Smirnov et al., 2021), where vehicle behaviors are largely constrained by upstream traffic conditions and the scope of interaction is relatively limited. In such scenarios, the advantage of intent-level reasoning is less pronounced, as cooperative negotiation among agents plays a minor role, although reservation-based or cooperative mechanisms have been explored in congestion mitigation (Yag\"{u}e-Cuevas et al., 2025).

Therefore, we select the traffic merging scenario as a representative complementary evaluation case to validate the generalization ability of the proposed LLM-based interactive decision-making framework.

\begin{figure*}[!t]
\centering
\subfloat[]{\includegraphics[width=0.48\textwidth]{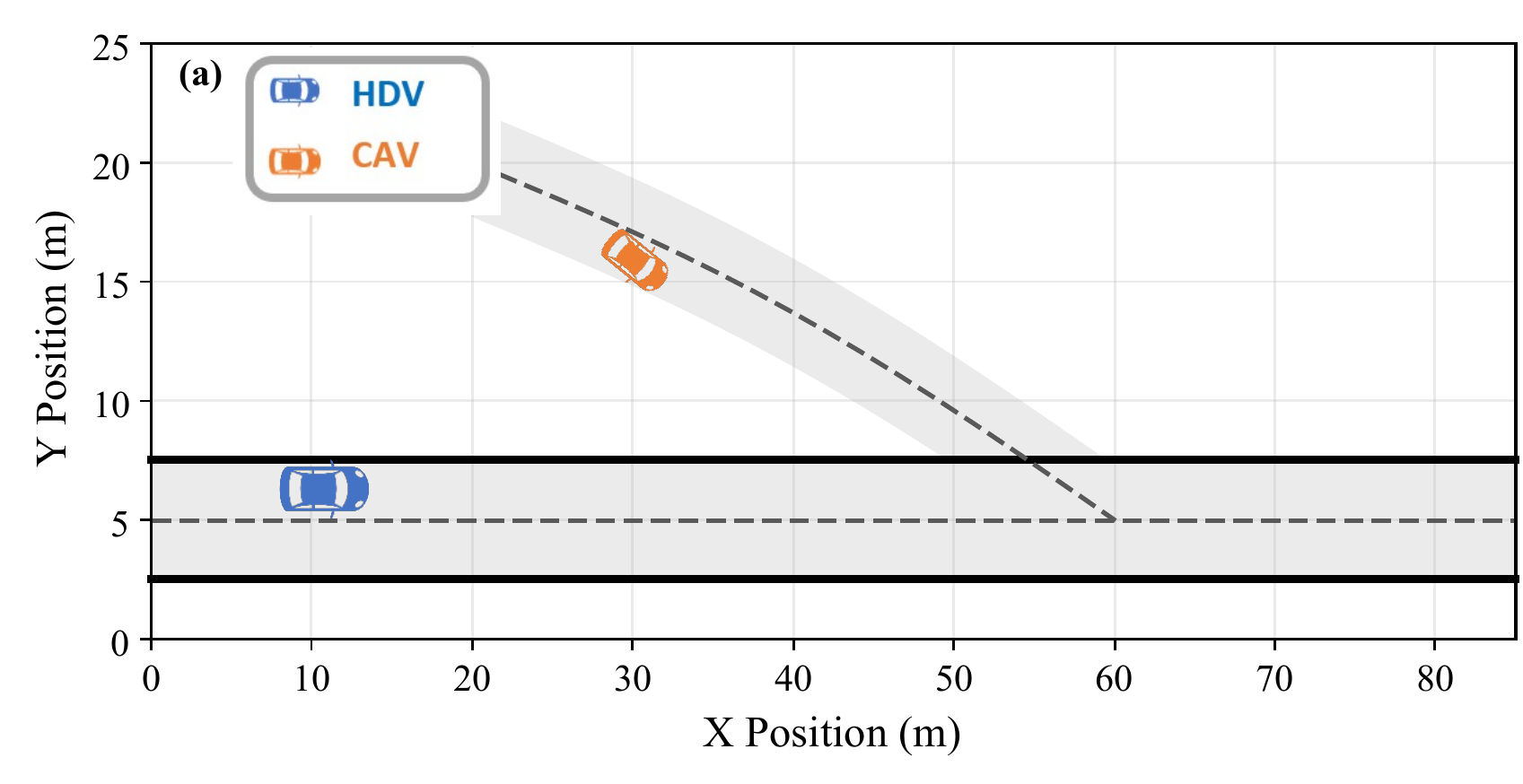}}
\hfill
\subfloat[]{\includegraphics[width=0.48\textwidth]{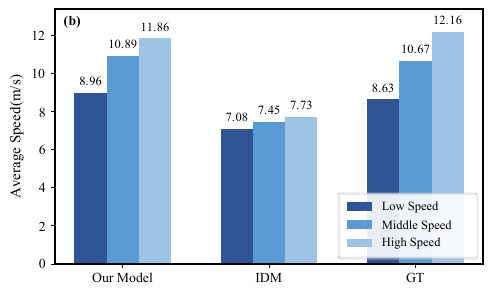}}
\caption{Comparison of OPM with other methods in merging scenario. (a) Experimental scenario. (b) Average speed across models.}
\label{fig:fig7}
\end{figure*}

Fig. 7(a) shows the experimental scenario, Fig. 7(b) shows the average speed results under different initial speed settings across models, the proposed method consistently achieves higher driving efficiency than the rule-based IDM model and remains comparable to the game-theoretic approach. Specifically, under low, medium, and high initial speed conditions, our method attains average speeds which are significantly higher than those of IDM and closely match the performance of the game-theoretic model.

These results indicate that, in interaction-intensive merging scenarios, the proposed LLM-based decision-making framework is able to effectively avoid overly conservative behaviors while preserving competitive driving efficiency. This suggests that the intent-level reasoning enabled by the proposed framework is not limited to the unprotected intersection case, but can also function in other structured interaction-critical scenarios with similar negotiation characteristics. However, the current evidence is still limited to representative simulator-based scenarios and should not be interpreted as sufficient validation for broader real-world generalization.

In conclusion, the interactive decision-making model based on LLM proposed in this paper has significantly improved traffic efficiency compared to traditional models, significantly enhancing driving efficiency; in terms of passenger comfort, it achieves smooth decision-making by combining actual interaction states, optimizing the driving experience; in terms of driving safety, through intent sharing and trajectory optimization technologies, it demonstrates stronger adaptability and comprehensive decision-making ability in various scenarios. These results fully demonstrate the practical potential and advantages of this research in the human-machine mixed driving environment.

\subsection{Interactive Autonomous Driving Decision Model Human-like Behavior Test}

To evaluate the human-like behavior of the proposed LLM-based interactive decision-making model, a Turing test was conducted using the SILAB driving simulator.

\begin{figure*}[!t]
\centering
\includegraphics[width=\textwidth]{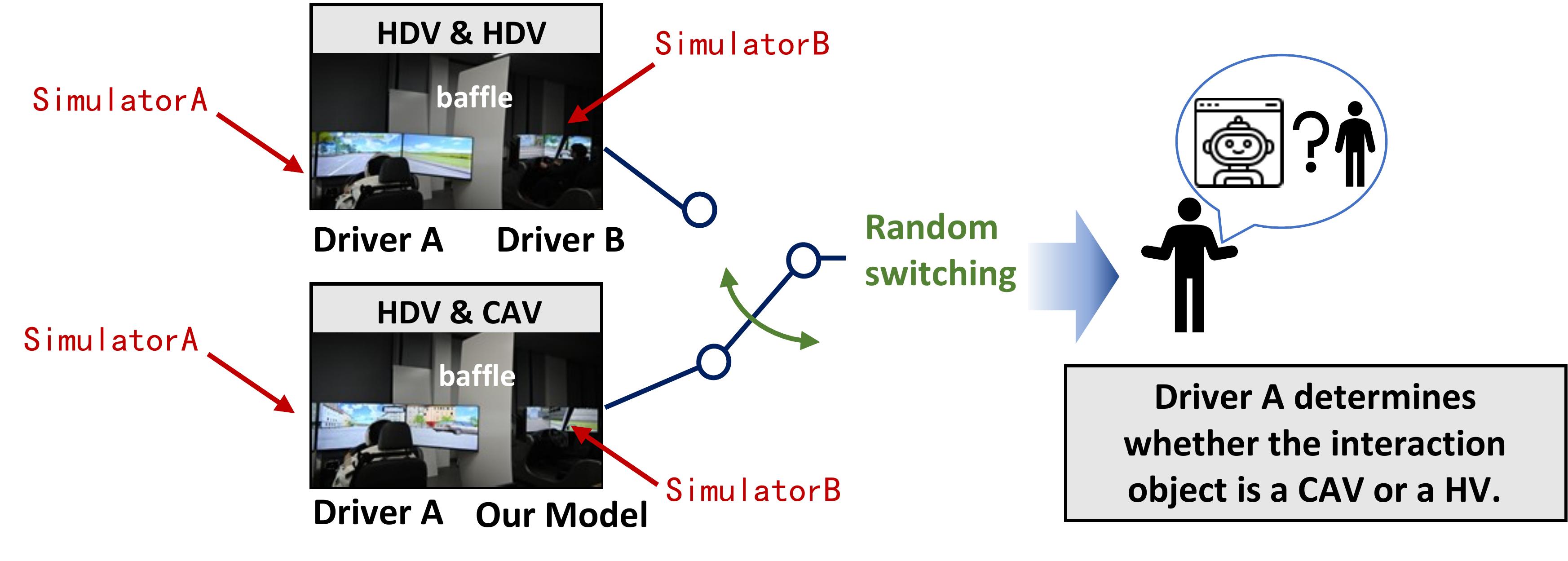}
\caption{Turing Test Experimental Scheme}
\label{fig:fig8}
\end{figure*}

The experimental scheme is shown in Fig. 8. The specific experimental process is as follows:

1) Scenario Setup: This experiment used two SILAB simulators connected online. Through their cluster driving function, a typical strong interactive scenario of left turn and straight ahead for two vehicles was constructed at a typical unsignalized intersection. The setup method was similar to that in Section 5.1. Driving simulator A controlled the vehicle\textquotesingle s driving path as straight ahead, with the initial position coordinates being (-40m, -5m), and was controlled by human driver A. Driving simulator B controlled the vehicle\textquotesingle s driving path as a left turn, with the initial position coordinates being (40m, 5m), and was controlled by human driver B or the LLM model in this paper. The two simulators were separated by a curtain wall to ensure that Driver A remained unaware of whether Simulator B was controlled by a human or the model during the experiment, thereby maintaining the double-blind nature of the Turing test.

2) Preparatory Stage: A total of 22 volunteers were recruited for this experiment, divided into groups A and B to control driving simulator A and B respectively. Each volunteer participated in two tests, collecting a total of 44 pairs of valid test data. Each participant was given 10 minutes of practice time before the formal test to become familiar with the simulator operation, and the control identity of Simulator B was determined only after seating to ensure that Driver A remained unaware of whether the interacting vehicle was controlled by a human driver or by the proposed model. This setting was intended to reduce prior expectation bias and preserve the blind nature of the evaluation.

3) Test Process: During the test, driving simulator A controlled the vehicle\textquotesingle s driving path as straight ahead, and human driver A decided whether to make an aggressive or conservative decision. Driving simulator B controlled the vehicle to make a left turn, and the decision was made autonomously by human driver B or the autonomous driving model. After both vehicles passed the intersection, the test task ended. After the test, human driver A needed to make a subjective judgment and score on the object controlling the vehicle in driving simulator B based on the interaction experience: determining whether it was human driving or an autonomous driving system, and the confidence level of this judgment was rated from 1 to 5.

The questionnaire results of the test were analyzed, as shown in Fig. 9. Regarding the accuracy rate, by looking at the confusion matrix in the Fig. 9, the overall accuracy rate was 0.57; within our sample size, this did not differ meaningfully from random guessing. This suggests that the interaction behavior generated by the proposed model exhibits a certain degree of similarity to human driving behavior, but the result should be interpreted as preliminary rather than definitive, given the current sample size and scenario coverage. The degree of human-likeness of the autonomous driving decision model proposed in this paper is already quite close to that of real human drivers. For the confidence level, the average value was 3.6/5, indicating that the subjects were not completely certain whether the other party was a human when making decisions, suggesting that the scheme in this paper caused the subjects to have a certain degree of confusion. For the naturalness of the other vehicle, the average score of the LLM model-controlled CAV was 3.9/5, while that of the human driver was 4.1/5, with a small difference, indicating that this scheme performed excellently in terms of behavioral naturalness and had the ability to approach human behavior.

\begin{figure}[!t]
\centering
\includegraphics[width=\linewidth]{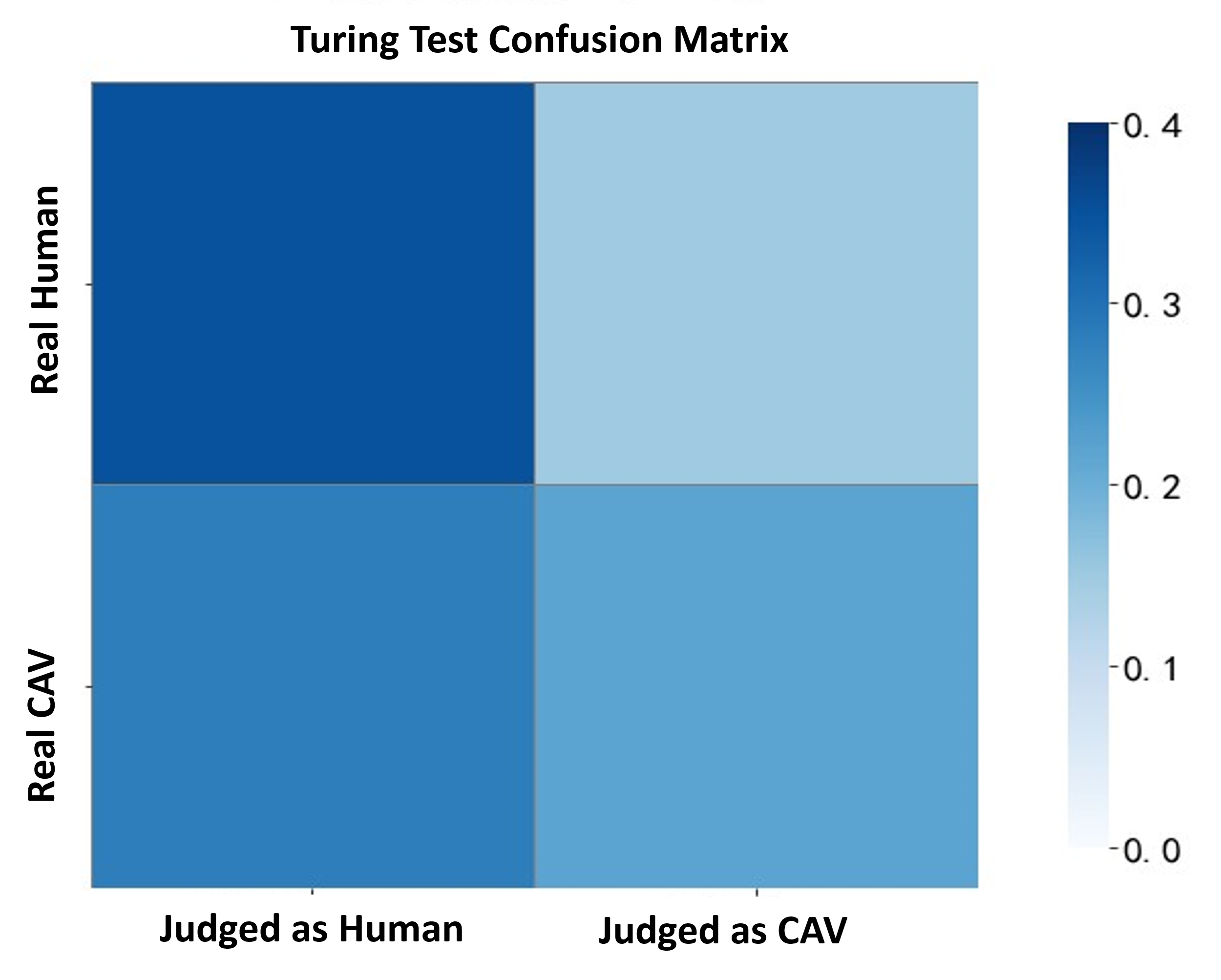}
\caption{Turing Test Confusion Matrix}
\end{figure}

In conclusion, the results of the Turing test experiment verified that the proposed LLM interactive decision-making model has good anthropomorphic characteristics in dynamic interaction scenarios. The model can make human-like interactive decisions based on the changes in the complex dynamic interaction scenarios and has the ability to interact with heterogeneous human drivers, which is conducive to enhancing the social acceptance of autonomous vehicles by human traffic participants.

\section{Conclusion}

This paper proposes an interactive autonomous driving decision-making method based on LLM, aiming to address the issues of conservative decision-making and insufficient interaction ability of autonomous vehicles in human-machine mixed driving environments. By introducing OPM semantic scene modeling, LLM intent parsing, Monte Carlo trajectory optimization, and eHMI natural language interaction, a closed-loop design from scene understanding to decision execution and interaction feedback is achieved. To verify the effectiveness of the proposed interactive autonomous driving decision-making, comprehensive performance comparison tests and Turing tests of different autonomous driving decision-making algorithms were conducted. The experimental results show that in the strong interaction scenario of unprotected left turn at an unsignalized intersection, the proposed model outperforms mainstream decision-making models in terms of safety, comfort, and efficiency indicators, and the Turing test results indicate that the model's degree of human-likeness approaches that of human drivers. However, these findings should still be interpreted within the scope of the current simulator-based evaluation rather than as evidence of broad real-world generalization.

The research has verified the feasibility and advantages of LLM in autonomous driving decision-making and interaction, providing a new path for improving the adaptability, safety, and social acceptance of autonomous vehicles in complex human-machine mixed driving environments. The current study focuses on validating the effectiveness of structured semantic scene abstraction and LLM-based interactive reasoning within a unified framework. More advanced reasoning strategies, such as chain-of-thought prompting and task-specific fine-tuning, may further improve semantic robustness and failure-case handling, but are beyond the present experimental scope. Future work will further explore the deep integration of multimodal perception and LLM, personalized adaptation strategies for intent interaction, and real vehicle testing and verification in real road environments by another journal.

\end{document}